\documentclass{article}
\usepackage{spconf,amsmath,graphicx}

\usepackage{booktabs}
\usepackage{amsmath,amssymb,amsthm}
\usepackage{algorithm}
\usepackage{algorithmic}
\usepackage[colorlinks,
            linkcolor=red, 
            anchorcolor=black,
            citecolor=green,
            ]{hyperref}

\title{EnsembleMOT: A Step towards Ensemble Learning of Multiple Object Tracking}
\name{Yunhao Du$^1$, Zihang Liu$^1$, Fei Su$^{1,2}$}
\address{
  $^1$ Beijing University of Posts and Telecommunications \\
  $^2$ Beijing Key Laboratory of Network System and Network Culture, China \\
  {\tt\small \{dyh\_bupt,henry0820,sufei\}@bupt.edu.cn}
}
\begin{document}
%
\maketitle
\begin{abstract}
  Multiple Object Tracking (MOT) has rapidly progressed in recent years.
  Existing works tend to design a single tracking algorithm to perform both detection and association.
  Though ensemble learning has been exploited in many tasks, i.e, classification and object detection,
  it hasn't been studied in the MOT task,
  which is mainly caused by its complexity and evaluation metrics.
  In this paper, we propose a simple but effective ensemble method for MOT, called \textbf{EnsembleMOT},
  which merges multiple tracking results from various trackers with spatio-temporal constraints.
  Meanwhile, several post-processing procedures are applied to filter out abnormal results.
  Our method is model-independent and doesn't need the learning procedure.
  What's more, it can easily work in conjunction with other algorithms, e.g., tracklets interpolation.
  Experiments on the MOT17 dataset demonstrate the effectiveness of the proposed method.
  Codes are available at \url{https://github.com/dyhBUPT/EnsembleMOT}.
\end{abstract}

\begin{figure*}[t]
  \centering
  \includegraphics[width = 1\textwidth]{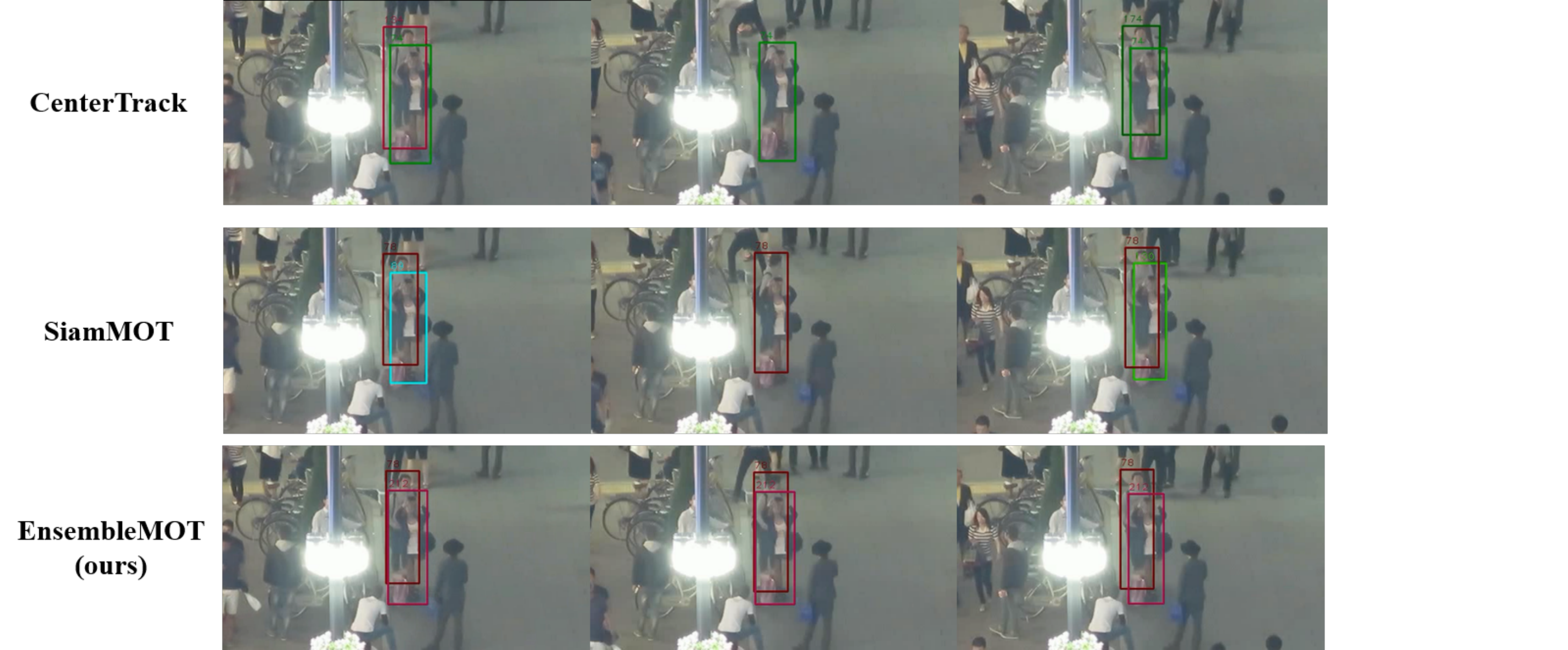}
  \caption{
    Example tracking results of CenterTrack \cite{zhou2020tracking}, SiamMOT \cite{shuai2021siammot} and EnsembleMOT(CenterTrack+SiamMOT).
    The same box color represents the same ID.
    The proposed EnsembleMOT algorithm obtains the best performances and produces more complete trajectories.
  }
  \label{fig_vis}
\end{figure*}

\section{Introduction}
\label{sec:intro}

Multiple Object Tracking (MOT) aims to detect and track all specific classes of objects frame by frame,
which plays an essential role in video analysis and understanding.
In the past few years, the MOT task is dominated by the tracking-by-detection (TBD) paradigm \cite{wojke2017simple, du2022strongsort},
which performs detection per frame and formulates the MOT problem as a data association task.
Recently, some works integrate the detector and embedding model (i.e., appearance or motion embedding) into a unified framework,
which can benefit from multi-task learning and tend to achieve a better speed-accuracy trade-off \cite{zhou2020tracking, zhang2021fairmot}.

Ensemble learning \cite{sagi2018ensemble} generally refers to training and/or combining multiple models,
which is widely used in machine learning \cite{efron1992bootstrap, breiman1996bagging, schapire1990strength, breiman1996stacked} 
and computer vision \cite{wortsman2022model, laine2016temporal, bodla1704improving, solovyev2021weighted}.
For example, for image classification, Wortsman et al. proposes Model Soups to average weights of multiple models to improve the classification accuracy \cite{wortsman2022model}.
To estimate more stable and accurate pseudo labels for semi-supervised image classification,
Temporal Ensembling \cite{laine2016temporal} aggregates the predictions of multiple previous network evaluations into an ensemble prediction.
For the object detection task, Soft-NMS \cite{bodla1704improving} and WBF \cite{solovyev2021weighted} are widely used to combine results from multiple detectors.

Ensemble methods are also used in several MOT works.
Peng et al. proposes the Layer-wise Aggregation Discriminative Model (LADM) \cite{peng2020dense},
which uses the weighted average of predictions from three softmax layers to judge whether a detection box represents a person or not.
However, it works in the detection procedure, and is essentially not for the tracking algorithm.
Inspired by SoftNMS, TrackNMS is designed in GIAOTracker \cite{du2021giaotracker} to fuse multiple tracking results.
It first sorts trajectories by the average confidence scores, and then performs non-maximum suppression (NMS) based on the temporal IoU.
Though it is designed for combining multiple trackers, it is evaluated by the score-based metrics mAP \cite{ILSVRC15},
in which redundant low-score results can benefit performance.
Instead, the instance-based metrics, i.e., MOTA \cite{bernardin2008evaluating}, IDF1\cite{ristani2016performance} and HOTA \cite{luiten2021hota},
are more common and reasonable evaluation metrics for the MOT task.

To sum up, ensemble methods used in the MOT task are still not well exploited.
We summarize the reasons as following:
\begin{itemize}
  \item MOT is a complex downstream task. 
  The diversity and complexity of various tracking algorithms makes it difficult to design a general and effective ensemble algorithm.
  \item The tracking results are temporal sequences, not just classification scores or detection bounding boxes (bboxes).
  Therefore, intuitive methods like voting can't be directly applied.
  \item The widely used metrics are instance-based.
  Compared with score-based metrics (e.g., mAP) in image classification and object detection,
  the instance-based metrics have no tolerance for redundant results,
  which introduces greater risk to ensemble methods.
\end{itemize}

In this paper, we propose a simple but effective ensemble method for instance-based metrics in the MOT task, called EnsembleMOT.
It mixes tracking results from multiple trackers, and then merges them based on spatio-temporal constraints.
The merged results may contain many redundant trajectories, so two post-processing methods,
bbox-level length-based NMS and sequence-level length-based filtering, are applied to refine the results.
Note that the proposed EnsembleMOT algorithm is model-independent and training-free,
which gives it great flexibility in real applications.
It doesn't need the training/testing data, 
which just takes multiple tracking results as input, and then outputs the ensemble results with significant improvements.

We conduct experiments on the MOT17 dataset \cite{milan2016mot16} with multiple state-of-the-art (SOTA) trackers as baseline,
i.e., SiamMOT\cite{shuai2021siammot}, CenterTrack\cite{zhou2020tracking}, TransTrack\cite{sun2020transtrack}, and FairMOT\cite{zhang2021fairmot}.
Results show that EnsembleMOT can improve metrics MOTA/IDF1/HOTA by 1 to 4.
Moreover, it can be integrated with other post-processing algorithms, e.g., interpolation \cite{du2022strongsort}.
To the best of our knowledge, it is the first ensemble algorithm for MOT with instance-based metrics.

\begin{algorithm}[t]
  \label{alg}
  \renewcommand{\algorithmicrequire}{\textbf{Input:}}  
	\renewcommand{\algorithmicensure}{\textbf{Output:}}  
	\caption{Merging procedure of EnsembleMOT} 
	\begin{algorithmic}[1]
		\REQUIRE Mixed multiple tracking results $\mathbb T = \{T_i\}_{i=1}^N$, 
    where $N$ is the number of trajectories and $T_i$ is one trajectory $T_i = \{b_i^t\}_{t=t_i^0}^{t=t_i^1}$;
    spatial threshold $thr_s$; temporal threshold $thr_t$.
		\ENSURE Ensemble tracking results $\mathbb T' = \{T'_j\}_{j=1}^M$, where $M$ is the number of resulting trajectories.
    \STATE $\mathbb T \gets sort(\mathbb T, key=len)$
    \STATE $\hat {\mathbb T} \gets \emptyset$ \ \ /* the merged trajectories */
    \STATE $\mathbb T' \gets \emptyset$ \ /* the final trajectories */
    \FOR{$T_i$ in $\mathbb T$}
      \IF{$T_i \in \hat {\mathbb T}$}
        \STATE $continue$
      \ENDIF
      \STATE $\tilde {\mathbb T} \gets \emptyset$ \ /* trajectories to be merged */
      \FOR{$T_j$ in $\mathbb T$ and $T_j.len < T_i.len$}
        \IF{$T_j \in \hat {\mathbb T}$}
          \STATE $continue$
        \ENDIF
        \STATE $stiou_{i,j} = stIoU(T_i, T_j, thr_s)$
        \IF{$stiou_{i,j} > thr_t$}
          \STATE $\tilde {\mathbb T} \gets \tilde {\mathbb T} \cup \{ T_j \}$
          \STATE $\hat {\mathbb T} \gets \hat {\mathbb T} \cup \{ T_j \}$
        \ENDIF
      \ENDFOR
      \STATE $\tilde T_i \gets merge(\tilde {\mathbb T}) $
      \STATE $\mathbb T' \gets \mathbb T' \cup \{ \tilde T_i \}$
    \ENDFOR
	\end{algorithmic} 
\end{algorithm}

\section{Method}
\label{sec:method}


\subsection{Spatio-Temporal IoU}

We utilize the spatio-temporal IoU (st-IoU) between trajectories as the constraints to merge them.
The trajectory $T_i$ is represented as a sequence of bounding boxes:
\begin{equation}
  T_i = \{ b_i^t \}_{t=t_i^0}^{t=t_i^1},
\end{equation}
where $b_i^t = [x_i^t, y_i^t, w_i^t, h_i^t]$ is the position of the bounding box of $T_i$ at frame $t$,
and $t_i^0, t_i^1$ are the start-stop frame id.

Given two trajectories $\{T_i, t_i^0, t_i^1\}$ and $\{T_j, t_j^0, t_j^1\}$ with partial overlaps,
i.e., $t_i^0 \le t_j^0 \le t_i^1 \le t_j^1$,
the frame-level spatial IoU is calculated at overlapping frames:
\begin{equation}
  sIoU_{i,j} = \{ IoU(b_i^t, b_j^t) \}_{t=t_j^0}^{t=t_i^1},
\end{equation}
where $IoU(\cdot, \cdot)$ is the intersection-over-union between two bounding boxes.

Then the temporal intersection between $T_i$ and $T_j$ is $inter_{i,j} = |sIoU'_{i,j}|$,
where
\begin{equation}
  sIoU'_{i,j} = \{ iou > Thr_s | iou \in sIoU_{i,j} \}
\end{equation}
is the subset of $sIoU_{i,j}$ whose elements are greater than spatial threshold $Thr_s$.
Similarly, the temporal union is $union_{i,j} = t_j^1 - t_i^0$.
Then, the st-IoU between $T_i$ and $T_j$ is calculated by
\begin{equation}
  stIoU_{i,j} = {inter_{i,j} \over union_{i,j}}. \label{stiou}
\end{equation}
For those trajectories with no overlaps, the st-IoU is set to 0.

Considering that when the two trajectories have a large length gap, the st-IoU would be small even if the shorter one is completely covered.
To solve this unreasonable case, we modify the denominator of Eq.(\ref{stiou}) to the length of the shorter trajectory,
i.e., $union_{i,j} = min(t_i^1 - t_i^0, t_j^1 - t_j^0)$.

\subsection{EnsembleMOT}
\label{subsec_ensemblemot}

The overall EnsembleMOT algorithm contains three steps, i.e., st-IoU-based mixture, bbox-level NMS and sequence-level filtering.

In the st-IoU-based mixture step, we assume the longer trajectories tend to be more reliable like \cite{du2021giaotracker}.
All trajectories are sorted in descending order of length, and then they are cycled through and compared with shorter trajectories.
In each loop, given the longer trajectory $T_i$, every shorter trajectory $T_j$ that has a st-IoU $stIoU_{i,j}$ larger than threshold $Thr_t$ with $T_i$
is considered to be the same ID with $T_i$.
Therefore, they are merged into one trajectory.
That is, for the overlapping frames, the final bboxes are the integration of corresponding bboxes from the two trajectories;
for the non-overlapping frames, the final bboxes are from the one that appears at that frame.
More details are listed in algorithm \ref{alg}.

The merged results would contain a lot of redundant bboxes that belongs to the same object.
To solve this problem, in the bbox-level NMS step, the non-maximum suppression with threshold $Thr_{nms}$ is applied to bboxes in each frame.
The only difference between it and standard NMS is,
the standard one uses confidence scores as the sorting criterion but the proposed method uses the length of trajectories.

In the followed sequence-level filtering step, trajectories shorter than $Thr_{len}$ is simply discarded
to remove those inaccuracy caused by short trajectories.

\subsection{Discussion}

The proposed EnsembleMOT algorithm doesn't need training process, codes of trackers or even datasets.
It only takes multiple tracking results as input and outputs more accurate results,
which brings great flexibility in applications.

Figure \ref{fig_vis} shows example tracking results of CenterTrack \cite{zhou2020tracking}, SiamMOT \cite{shuai2021siammot} and EnsembleMOT (CenterTrack + SiamMOT).
Just bboxes of two objects (a woman and a man) are visualized for clarity.
For CenterTrack, the woman (in green bbox) is well tracked, but the man has an ID switch.
However, for SiamMOT, the man (in rufous bbox) is well tracked, but not for the woman.
The third row presents the ensemble results of CenterTrack and SiamMOT, in which both the man and the woman are well tracked.
In other words, the EnsembleMOT algorithm can obtain the best results of both trackers and produce more complete trajectories. 

Strictly speaking, EnsembleMOT is an ensemble algorithm rather than an ensemble learning algorithm, because it doesn't need learning.
However, we hope that it can serve as a simple baseline and inspire more works on ensemble learning in MOT.

\begin{table*}[t]
  \begin{center}
    \caption{
      Results of applying EnsembleMOT and GSI on various SOTA trackers on the MOT17 dataset.
    }
    \label{table_ablation1}
    \resizebox{0.9\textwidth}{!}{
      \begin{tabular}{cl|c|c|l|l|l}
        \toprule[1pt]
        & \textbf{Tracker(s)} & \textbf{EnsembleMOT} & \textbf{GSI} & \textbf{MOTA(↑)} & \textbf{IDF1(↑)} & \textbf{HOTA(↑)} \\
        \hline
        & SiamMOT     \cite{shuai2021siammot}  &     -      &     -      & 62.32 & 67.35 & 56.52 \\
        & CenterTrack \cite{zhou2020tracking}  &     -      &     -      & 66.80 & 64.45 & 55.30 \\
        & TransTrack  \cite{sun2020transtrack} &     -      &     -      & 67.72 & 68.59 & 58.09 \\
        & FairMOT     \cite{zhang2021fairmot}  &     -      &     -      & 69.14 & 72.66 & 57.32 \\
        \hline
        & FairMOT + SiamMOT                    & \checkmark &     -      & 71.57 (\textbf{+2.43}) & 74.53 (+1.87) & 59.84 (+2.52) \\
        &                                      & \checkmark & \checkmark & 71.41 (+2.27) & 74.61 (\textbf{+1.95}) & 60.04 (\textbf{+2.72}) \\
        \hline
        & FairMOT + TransTrack                 & \checkmark &     -      & 71.61 (\textbf{+2.47}) & 74.20 (+1.54) & 60.19 (+2.10) \\
        &                                      & \checkmark & \checkmark & 71.25 (+2.11) & 74,67 (\textbf{+2.01}) & 60.82 (\textbf{+2.73}) \\
        \hline
        & FairMOT + CenterTrack                & \checkmark &     -      & 71.53 (+2.39) & 74.41 (+1.75) & 59.97 (+2.65) \\
        &                                      & \checkmark & \checkmark & 72.59 (\textbf{+3.45}) & 75.23 (\textbf{+2.57}) & 61.06 (\textbf{+3.74}) \\
        \hline
        & SiamMOT + CenterTrack                & \checkmark &     -      & 67.72 (\textbf{+0.92}) & 71.66 (\textbf{+4.31}) & 59.67 (\textbf{+3.15}) \\
        &                                      & \checkmark & \checkmark & 67.00 (+0.20) & 71.39 (+4.04) & 59.59 (+3.07) \\
        \hline
        & TransTrack + CenterTrack             & \checkmark &     -      & 68.47 (+0.75) & 70.03 (+1.44) & 58.85 (+0.76) \\
        &                                      & \checkmark & \checkmark & 69.22 (\textbf{+1.50}) & 70.46 (\textbf{+1.87}) & 59.42 (\textbf{+1.33}) \\
        \hline
        & SiamMOT + TransTrack                 & \checkmark &     -      & 68.42 (\textbf{+0.70}) & 71.92 (\textbf{+3.33}) & 60.17 (\textbf{+2.08}) \\
        &                                      & \checkmark & \checkmark & 67.69 (-0.03) & 71.57 (+2.98) & 60.06 (+1.97) \\
        \bottomrule[1pt]
      \end{tabular}
    }

    \caption{
      Ablation results of different components of EnsembleMOT based on "SiamMOT+CenterTrack".
    }
    \label{table_ablation2}
    \resizebox{0.9\textwidth}{!}{
      \begin{tabular}{cl|c|c|c|c|c|c}
        \toprule[1pt]
        & \textbf{Trackers} & \textbf{dropping} & \textbf{NMS} & \textbf{Filtering} & \textbf{MOTA(↑)} & \textbf{IDF1(↑)} & \textbf{HOTA(↑)} \\
        \hline
        & SiamMOT               &            &            &            & 62.32          & 67.35          & 56.52 \\
        & CenterTrack           &            &            &            & 66.80          & 64.45          & 55.30 \\
        & SiamMOT + CenterTrack &            &            &            & 66.28          & 70.39          & 59.07 \\
        &                       & \checkmark &            &            & 67.17          & 70.71          & 59.35 \\
        &                       & \checkmark & \checkmark &            & \textbf{67.73} & 70.84          & 59.38 \\
        &                       & \checkmark & \checkmark & \checkmark & 67.72          & \textbf{71.66} & \textbf{59.67} \\
        \bottomrule[1pt]
      \end{tabular}
    }
  \end{center}
\end{table*}

\section{Experiments}
\label{sec:exper}

\subsection{Datasets and Evaluation Metrics}

\noindent \textbf{Datasets.}
To verify the effectiveness of the proposed method, we conduct experiments on the MOT17 \cite{milan2016mot16} datasets.
MOT17 is a popular dataset for MOT, which consists of 7 sequences, 5,316 frames for training and 7 sequences, 5919 frames for testing.
For ablation studies, recent works generally take the first half of each sequence in the MOT17 training set for training 
and the last half for validation.
Following them, we conduct experiments on the validation set.

\noindent \textbf{Metrics.}
We use the metrics MOTA, IDF1 and HOTA
to evaluate tracking performance \cite{bernardin2008evaluating, ristani2016performance, luiten2021hota}.
MOTA is computed based on FP, FN and IDs, and focuses more on detection performance.
By comparison, IDF1 better measures the consistency of ID matching. 
HOTA is an explicit combination of detection score DetA and association score AssA, 
which balances the effects of performing accurate detection and association into a single unified metric.

\subsection{Implementation Details}

We select four recent SOTA trackers as our baseline methods, 
i.e., SiamMOT\cite{shuai2021siammot}, CenterTrack\cite{zhou2020tracking}, TransTrack\cite{sun2020transtrack}, and FairMOT\cite{zhang2021fairmot}.
We run the official codes of these trackers with default settings to reproduce their results.
For EnsembleMOT, in all experiments, 
we set spatial threshold $Thr_s = 0.5$, temporal threshold $Thr_t = 0.5$, NMS threshold $Thr_{nms}=0.7$ and length threshold $Thr_{len} = 20$.
For GSI, we use the default settings in \cite{du2022strongsort}.

\subsection{Main Results}

Table \ref{table_ablation1} presents the comparison between baseline trackers and our EnsembleMOT.
It's shown that EnsembleMOT can improve the metrics by a large margin.
Moreover, it can also work in conjunction with the interpolation algorithm, GSI \cite{du2022strongsort}.
Specifically, compared with FairMOT, 
the ensemble results of "FairMOT+CenterTrack" improves its MOTA by 2.39, IDF1 by 1.75, HOTA by 2.65.
By applying GSI, the improvements become 3.45, 2.57 and 3.74.

\subsection{Ablations}

Table \ref{table_ablation2} presents the ablation results of different components of EnsembleMOT.
We use SiamMOT and CenterTrack as the basic trackers (in the first and second row).
The baseline ensemble method in the third row only applies the merging procedure 
which utilizes the averaging bboxes from the two trajectories as the integrated bboxes in overlapping frames.
The "dropping" means reserving the bbox from the longer trajectory and dropping other bboxes, rather than averaging them.
This can reduce the cost of false merging.
The "NMS" represents bbox-level NMS and "Filtering" represents sequence-level filtering.
It is obvious that they can refine the detection and association results respectively.

\section{Conclusion}
\label{sec:conc}

In this paper, we propose a simple but effective ensemble algorithm for the MOT task, EnsembleMOT.
It doesn't need learning and has great flexibility in applications.
Experiments demonstrate its effectiveness over four SOTA trackers.
It can also work together with other post-processing methods, i.e., tracklets interpolation.
We hope the proposed EnsembleMOT can serve as a simple baseline and inspire more ensemble works for MOT.
In future work, we will exploit more general learning-based ensemble methods.

\section{Acknowledgments}
This work is supported by Chinese National Natural Science Foundation under Grants (62076033, U1931202).

\vfill\pagebreak

\bibliographystyle{IEEEbib}
\bibliography{strings,refs}

\end{document}